\pdfoutput=1

\documentclass[11pt,a4paper]{article}
\usepackage[hyperref]{acl2021}
\usepackage{times}

\usepackage{svg}
\usepackage{latexsym}
\usepackage{booktabs}
\usepackage{setspace}
\usepackage{amsmath}
\usepackage{amssymb}
\usepackage{dsfont}
\usepackage{adjustbox}
\usepackage{array}
\usepackage{pgfplotstable}
\usepackage{graphics}
\usepackage{multirow,multicol}

\usepackage{microtype}

 \aclfinalcopy
\def\aclpaperid{3591}

\title{How to Adapt Your Pretrained Multilingual Model to 1600 Languages}

\author{Abteen Ebrahimi \and Katharina Kann \\
  University of Colorado Boulder \\
  \texttt{\{abteen.ebrahimi,katharina.kann\}@colorado.edu}}

\date{}

\begin{document}
\maketitle
\begin{abstract}
Pretrained multilingual models (PMMs) enable zero-shot learning via cross-lingual transfer, performing best for languages seen during pretraining. While methods exist to improve performance for unseen languages, they have almost exclusively been evaluated using amounts of raw text only available for a small fraction of the world's languages. In this paper, we evaluate the performance of existing methods to adapt PMMs to new languages using a resource available for over 1600 languages: the New Testament. This is challenging for two reasons: (1) the small corpus size, and (2) the narrow domain. While performance drops for all approaches, we surprisingly still see gains of up to $17.69\%$ accuracy for part-of-speech tagging and $6.29$ F1 for NER on average over all languages as compared to XLM-R. Another unexpected finding is that continued pretraining, the simplest approach, performs best. Finally, we perform a case study to disentangle the effects of domain and size and to shed light on the influence of the finetuning source language.

\end{abstract}

\section{Introduction}

Pretrained multilingual models (PMMs) are a straightforward way to enable zero-shot learning via cross-lingual transfer, thus eliminating the need for labeled data for the target task and language. However, downstream performance is highest for languages that are well represented in the pretraining data or linguistically similar to a well represented language. Performance degrades as representation decreases, with
languages not seen during pretraining generally having the worst performance. In the most extreme case, when a language's script is completely unknown to the model, zero-shot performance is effectively random.

While multiple methods have been shown to improve the performance of transfer to underrepresented languages (cf. Section \ref{methods}), previous work has evaluated them using unlabeled data from sources available for
a relatively small number of languages, such as Wikipedia or Common Crawl, which cover 316\footnote{\url{https://en.wikipedia.org/wiki/List_of_Wikipedias}} and 160\footnote{\url{https://commoncrawl.github.io/cc-crawl-statistics/plots/languages}} languages, respectively. Due to this low coverage, the languages that would most benefit from these methods are precisely those which do not have the necessary amounts of monolingual data to implement them as-is.
To enable the use of PMMs for truly low-resource languages, where they can, e.g., assist language documentation or revitalization, it is important to understand how state-of-the-art adaptation methods act in a setting more broadly applicable to many languages.

\begin{figure}[t]
    \centering
    \includegraphics[width=\columnwidth]{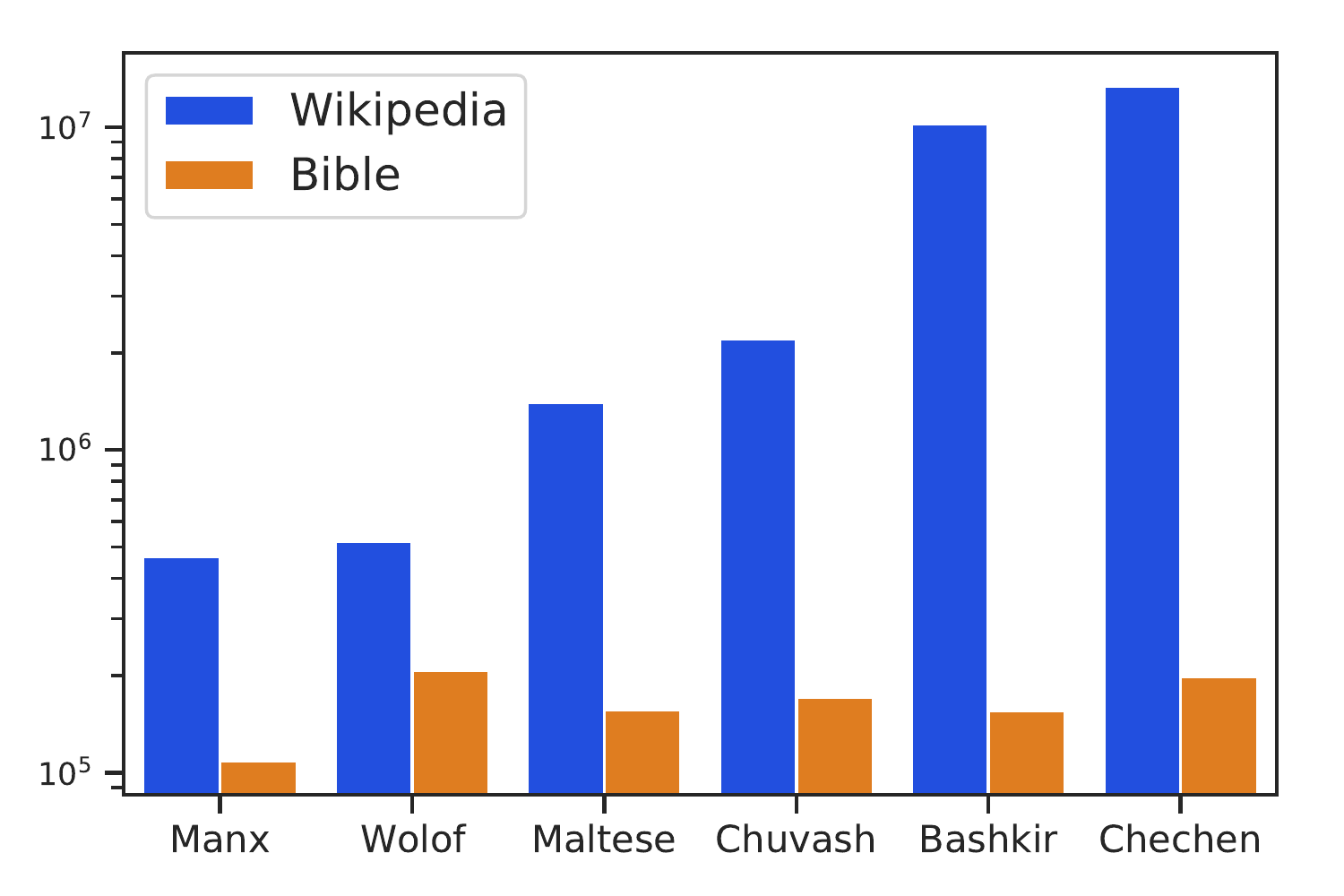}
    \caption{The number of space-separated words in the Bible and Wikipedia for six low-resource languages used in our experiments; plotted on a log scale. }
    \label{fig:fpfig}
\end{figure}

In this paper, we ask the following question: \emph{Can we use the Bible -- a resource available for roughly 1600 languages -- to improve a PMM's zero-shot performance on an unseen target language? And, if so, what adaptation method works best?}
We investigate the performance of XLM-R \cite{Conneau2020UnsupervisedCR} when combined with continued pretraining \cite{mbert-parsing}, vocabulary extension, \cite{wang-etal-2020-extending}, and adapters \cite{Pfeiffer2020MADXAA} making the following assumptions:
(1) the only text available in a target language is the New Testament,  and (2) no annotated training data exists in the target language.

We present results on 2 downstream tasks -- part-of-speech (POS) tagging and named entity recognition (NER) -- on a typologically diverse set of 30 languages, all of which are unseen during the pretraining of XLM-R.  We find that, surprisingly, even though we use a small corpus from a narrow domain, most adaptation approaches improve over XLM-R's base performance, showing that the Bible is a valuable source of data for our purposes. We further observe that in our setting the simplest adaptation method, continued pretraining, performs best for both tasks, achieving gains of up to 17.69\% accuracy for POS tagging, and 6.29 F1 for NER on average across languages.

Additionally, we seek to disentangle the effects of two aspects of our experiments on downstream performance: the selection of the source language, and the restricted domain of the New Testament. Towards this, we conduct a case study focusing on three languages with Cyrillic script: Bashkir, Chechen, and Chuvash. In order to understand the effect of the choice of source language, we use a more similar language, Russian, as our source of labeled data. To explore the effect of the New Testament's domain, we conduct our pretraining experiments with an equivalent amount of data sampled from the Wikipedia in each language.
We find that
changing the source language to Russian increases average baseline performance by 18.96 F1, and we achieve the highest results across all settings when using both Wikipedia and Russian data.

\begin{table}[]
    \centering
    \small
    \begin{adjustbox}{width=\columnwidth}
    \begin{tabular}{lllll}
        \toprule
        Code & Language & Script & Language Family & Task \\
        \toprule
        ace & Acehnese & Latin  & Austronesian & NER \\
        arz & Egyptian Arabic & Arabic  & Afro-Asiatic & NER \\
        bak & Bashkir & Cyrillic  & Turkic & NER \\
        bam & Bambara & Latin, N'ko  & Mande & POS \\
        ceb & Cebuano & Latin  & Austronesian & NER \\
        che & Chechen & Cyrillic  & Northeast Caucasian & NER \\
        chv & Chuvash & Cyrillic  & Turkic & NER \\
        cop & Coptic & Coptic  & Ancient Egyptian & POS \\
        crh & Crimean Turkish & Cyrillic  & Turkic & NER \\
        glv & Manx & Latin  & Indo-European & POS \\
        grc & Ancient Greek & Greek  & Indo-European & POS \\
        gsw & Swiss German & Latin  & Indo-European & POS \\
        hak & Hakka Chinese & Chinese  & Sino-Tibetan & NER \\
        ibo & Igbo & Latin   & Niger-Congo & NER \\
        ilo & Iloko & Latin  & Austronesian & NER \\
        kin & Kinyarwanda & Latin  & Niger-Congo & NER \\
        mag & Magahi & Devanagari  & Indo-Iranian & POS \\
        mhr & Eastern Mari & Cyrillic  & Uralic & NER \\
        min & Minangkabau & Latin  & Austronesian & NER \\
        mlt & Maltese & Latin  & Afro-Asiatic & Both \\
        mri & Maori & Latin  & Austronesian & NER \\
        myv & Erzya & Cyrillic  & Uralic & POS \\
        nds & Low German & Latin  & Indo-European & NER \\
        ory & Odia & Odia  & Indo-Iranian & NER \\
        sco & Scots & Latin  & Indo-European & NER \\
        tat & Tatar & Cyrillic  & Turkic & NER \\
        tgk & Tajik & Cyrillic  & Indo-Iranian & NER \\
        war & Waray & Latin  & Austronesian & NER \\
        wol & Wolof & Latin  & Niger-Congo & Both \\
        yor & Yoruba & Latin  & Niger-Congo & Both \\
    \bottomrule
    \end{tabular}
    \end{adjustbox}
    \caption{Languages used in our experiments, none of which are represented in XLM-R's pretraining data.}

    \label{tab:languages}
\end{table}

\section{Related Work}

\subsection{Background}

Prior to the introduction of PMMs, cross-lingual transfer was often based on word embeddings \cite{mikolov13}. \citet{joulin2018loss} present monolingual embeddings for 294 languages using Wikipedia, succeeded by \citet{grave2018learning} who present embeddings for 157 languages trained on additional data from Common Crawl. For cross-lingual transfer, monolingual embeddings can then be aligned using existing parallel resources, or in a completely unsupervised way \cite{bojanowski-etal-2017-enriching, artetxe-etal-2017-learning, lample2017unsupervised, conneau2017word,artetxe-etal-2016-learning}. Although they use transformer based models,  \citet{Artetxe2020OnTC} also transfer in a monolingual setting. Another method for cross-lingual transfer involves multilingual embeddings, where languages are jointly learned as opposed to being aligned \cite{Ammar2016MassivelyMW, Artetxe2019MassivelyMS}. For a more in-depth look at cross-lingual word embeddings, we refer the reader to \citet{Ruder_2019}.

While the above works deal with generally improving cross-lingual representations, task-specific cross-lingual systems often show strong performance in a zero-shot setting. For POS tagging, in a similar setting to our work, \citet{eskander-etal-2020-unsupervised} achieve strong zero-shot results by using unsupervised projection \cite{yarowsky-etal-2001-inducing} with aligned Bibles. Recent work for cross-lingual NER includes \citet{mayhew-etal-2017-cheap} who use dictionary translations to create target-language training data, as well as \citet{xie-etal-2018-neural} who use a bilingual dictionary in addition to self-attention. \citet{bharadwaj-etal-2016-phonologically} use phoneme conversion to aid cross-lingual NER in a zero-shot setting. More recently, \citet{BARI2020ZeroResourceCN} propose a model only using monolingual data for each language, and \citet{qi-etal-2020-stanza} propose a language-agnostic toolkit supporting NER for 66 languages. In contrast to these works, we focus on the improvements offered by adaptation methods for pretrained models for general tasks.

\subsection{Pretrained Multilingual Models}
PMMs can be seen as the natural extension of multilingual embeddings to pretrained transformer-based models.
mBERT was the first PMM, covering the 104 languages with the largest Wikipedias. It uses a 110k byte-pair encoding (BPE) vocabulary \cite{sennrich-etal-2016-neural} and is pretrained on both a next sentence prediction and a masked language modeling (MLM) objective. Languages with smaller Wikipedias are upsampled and highly represented languages are downsampled.
XLM is a PMM trained on 15 languages. XLM similarly trains on Wikipedia data, using a BPE vocabulary with 95k subwords and up- and downsamples languages similarly to mBERT. XLM also introduces translation language modeling (TLM), a supervised pretraining objective, where tokens are masked as for MLM, but parallel sentences are concatenated such that the model can rely on subwords in both languages for prediction. Finally, XLM-R is an improved version of XLM. Notable differences include the larger vocabulary of 250k subwords created using SentencePiece tokenization \cite{kudo-richardson-2018-sentencepiece} and the training data, which is taken from CommonCrawl and is considerably more than for mBERT and XLM. XLM-R relies solely on MLM for pretraining and achieves state-of-the-art results on multiple benchmarks \cite{Conneau2020UnsupervisedCR}.
We therefore focus solely on XLM-R in our experiments.

\paragraph{Downstream Performance of PMMs}
While \citet{pires-etal-2019-multilingual} and \citet{wu-dredze-2019-beto} show the strong zero-shot performance of mBERT, \citet{Wu2020AreAL} shine light on the difference in performance between well and poorly represented languages after finetuning on target-task data.

\citet{Muller2020WhenBU} observe varying zero-shot performance of mBERT on different languages not present in its pretraining data. They group them into `easy' languages, on which mBERT performs well without any modification, `medium' languages, on which mBERT performs well after additional pretraining on monolingual data, and `hard' languages, on which mBERT's performs poorly even after modification. They additionally note the importance of script, finding that transliterating into Latin offers improvements for some languages. As transliteration involves language specific tools, we consider it out of scope for this work, and leave further investigation in how to best utilize transliteration for future work. \citet{lauscher-etal-2020-zero} focus on PMM finetuning, and find that for unseen languages, gathering labeled data for few-shot learning may be more effective than gathering large amounts of unlabeled data.

Additionally, \citet{mbert-parsing}, \citet{wang-etal-2020-extending}, and \citet{Pfeiffer2020MADXAA} present the adaptation methods whose performance we investigate here in a setting where only the Bible is available. We give a general overview of these methods in the remainder of this section, before describing their application in our experiments in Section \ref{experiments}. 

\subsection{Adaptation Methods}
\label{methods}

\paragraph{Continued Pretraining} In a monolingual setting, continued pretraining of a language representation model on an MLM objective has shown to help downstream performance on tasks involving text from a domain distant from the pretraining corpora \cite{Gururangan2020DontSP}. In a multilingual setting, it has been
found that, given a target language, continued pretraining on monolingual data from that language can
lead to improvements on downstream tasks \cite{mbert-parsing, Muller2020WhenBU}.

\paragraph{Vocabulary Extension}
Many pretrained models make use of a subword vocabulary, which strongly reduces
the issue of out-of-vocabulary tokens.
However, when the pretraining and target-task domains differ, important domain-specific words may be over-fragmented, which reduces performance. In the monolingual setting, \citet{zhang-etal-2020-multi-stage} show that extending the vocabulary with in-domain tokens yields
performance gains. A similar result to that of continued pretraining holds in the multilingual setting: downstream performance of an underrepresented language benefits from additional tokens in the vocabulary, allowing for better representation of that language. \citet{wang-etal-2020-extending} find that extending the vocabulary of mBERT with new tokens and training on a monolingual corpus yields improvements for a target language, regardless of whether the language was seen or unseen. \citet{mbert-parsing} have similar results, and introduce tiered vocabulary augmentation, where new embeddings are learned with a higher learning rate. While both approaches start with a random initialization, they differ in the amount of new tokens added: \citet{wang-etal-2020-extending} limit new subwords to 30,000, while \citet{mbert-parsing} set a limit of 99, selecting the subwords which reduce the number of unknown tokens while keeping the subword-to-token ratio similar to the original vocabulary.

\paragraph{Adapters}
Adapters are layers with a small number of parameters, injected into models to help transfer learning \cite{adapter1}. \citet{adapter-houlsby} demonstrate the effectiveness of task-specific adapters in comparison to standard finetuning. \citet{Pfeiffer2020MADXAA} present invertible adapters and MAD-X, a framework utilizing them along with language and task adapters for cross-lingual transfer. After freezing model weights, invertible and language adapters for each language, including English, are trained together using MLM. The English-specific adapters are then used along with a task adapter to learn from labeled English data. For zero-shot transfer, the invertible and language adapters are replaced with those trained on the target language, and the model is subsequently evaluated.

\section{Experiments}
\label{experiments}

\subsection{Data and Languages}
\paragraph{Unlabeled Data}
We use the Johns Hopkins University Bible Corpus (JHUBC) from \citet{mccarthy-etal-2020-johns}, which contains 1611 languages, providing verse-aligned translations of both the Old and New Testament. However, the New Testament is much more widely translated: 86\% of translations do not include the Old Testament. We therefore
limit our experiments to the
New Testament, which accounts to about 8000 verses in total, although specific languages may not have translations of all verses. For the 30 languages we consider, this averages to around 402k subword tokens per language. The specific versions of the Bible we use are listed in Table \ref{tab:bible_counts}.

\paragraph{Labeled Data}
For NER, we use the splits of \citet{rahimi-etal-2019-massively}, which are created from the WikiAnn dataset \cite{pan-etal-2017-cross}. For POS tagging, we use data taken from the Universal Dependencies Project \cite{Nivre2020UniversalDV}. As XLM-R utilizes a subword vocabulary, we perform sequence labeling by assigning labels to the last subword token of each word. For all target languages, we only finetune on labeled data in English.

\paragraph{Language Selection } To select the languages for our experiments, we first compile lists of all languages for which a test set exists for either downstream task and we have a Bible for. We then filter these languages by removing those present in the pretraining data of XLM-R.
See Table \ref{tab:languages} for a summary of languages, their attributes, and the downstream task we use them for.

\subsection{PMM Adaptation Methods} 
Our goal is to analyze state-of-the-art PMM adaption approaches
in a true low-resource setting where the only raw text data available comes from the New Testament and no labeled data exists at all.
We now describe our implementation of these methods.
We focus on the \texttt{Base} version of \texttt{XLM-R} \cite{Conneau2020UnsupervisedCR} as our baseline PMM.

\paragraph{Continued Pretraining} We consider three models based on continued pretraining. In the simplest case, \texttt{+MLM}, we continue training XLM-R with an MLM objective on the available verses of the New Testament. Additionally, as Bible translations are a parallel corpus, we also consider a model, \texttt{+TLM}, trained using translation language modeling. Finally, following the findings of \citet{Lample2019CrosslingualLM}, we also consider a model using both TLM and MLM, \texttt{+\{M|T\}LM}. For this model, we alternate between batches consisting solely of verses from the target Bible and batches consisting of aligned verses of the target-language and source-language Bible. For NER, we pretrain \texttt{+MLM} and \texttt{+TLM} models for 40 epochs, and pretrain \texttt{+\{M|T\}LM} models for 20 epochs. For POS tagging, we follow a simlar pattern, training \texttt{+MLM} and \texttt{+TLM} for 80 epochs, and \texttt{+\{M|T\}LM} for 40 epochs. 

\paragraph{Vocabulary Extension} To extend the vocabulary of XLM-R, we implement the process of \citet{wang-etal-2020-extending}. We denote this as \texttt{+Extend}. For each target language, we train a new SentencePiece \cite{kudo-richardson-2018-sentencepiece} tokenizer on the Bible of that language with a maximum vocabulary size of 30,000.\footnote{We note that for many languages, the tokenizer cannot create the full 30,000 subwords due to the limited corpus size.} To prevent adding duplicates, we filter out any subword already present in the vocabulary of XLM-R. We then add additional pieces representing these new subwords into the tokenizer of XLM-R, and increase XLM-R's embedding matrix accordingly using a random initialization. Finally, we train the embeddings using MLM on the Bible. For NER, we train \texttt{+Extend} models for 40 epochs, and for POS tagging, we train for 80 epochs.  

\paragraph{Adapters} For adapters, we largely follow the full MAD-X framework \cite{Pfeiffer2020MADXAA}, using language, invertible, and task adapters. This is denoted as \texttt{+Adapters}. To train task adapters, we download language and invertible adapters for the source language from AdapterHub \cite{pfeiffer2020AdapterHub}. We train a single task adapter for each task, and use it across all languages. We train language and invertible adapters for each target language by training on the target Bible with an MLM objective. As before, for NER we train for 40 epochs, and for POS we train for 80 epochs.

\subsection{Hyperparameters and Training Details} 
\label{training-details}

For finetuning, we train using 1 Nvidia V100 32GB GPU, and use an additional GPU for adaptation methods. Experiments for NER and POS take around 1 and 2 hours respectively, totalling to 165 total training hours, and 21.38 kgCO$_2$eq emitted \cite{lacoste2019quantifying}. All experiments are run using the Huggingface Transformers library \cite{wolf-etal-2020-transformers}. We limit sequence lengths to 256 tokens.

We select initial hyperparameters for finetuning by using the English POS development set. We then fix all hyperparameters other than the number of epochs, which we tune using the 3 languages which have development sets, Ancient Greek, Maltese, and Wolof. We do not use early stopping. For our final results, we finetune for 5 epochs with a batch size of 32, and a learning rate of 2e-5. We use the same hyperparameters for both tasks. 

For each task and adaptation approach, we search over \{10, 20, 40, 80\} epochs, and select the epoch which gives the highest average performance across the development languages. We use the same languages as above for POS. For NER we use 4 languages with varying baseline performances: Bashkir, Kinyarwanda, Maltese, and Scots. We pretrain with a learning rate of 2e-5 and a batch size of 32, except for \texttt{+Adapters}, for which we use a learning rate of 1e-4 \cite{Pfeiffer2020MADXAA}.
\begin{figure*}[ht]
    \centering
    \includegraphics[width=.9\textwidth]{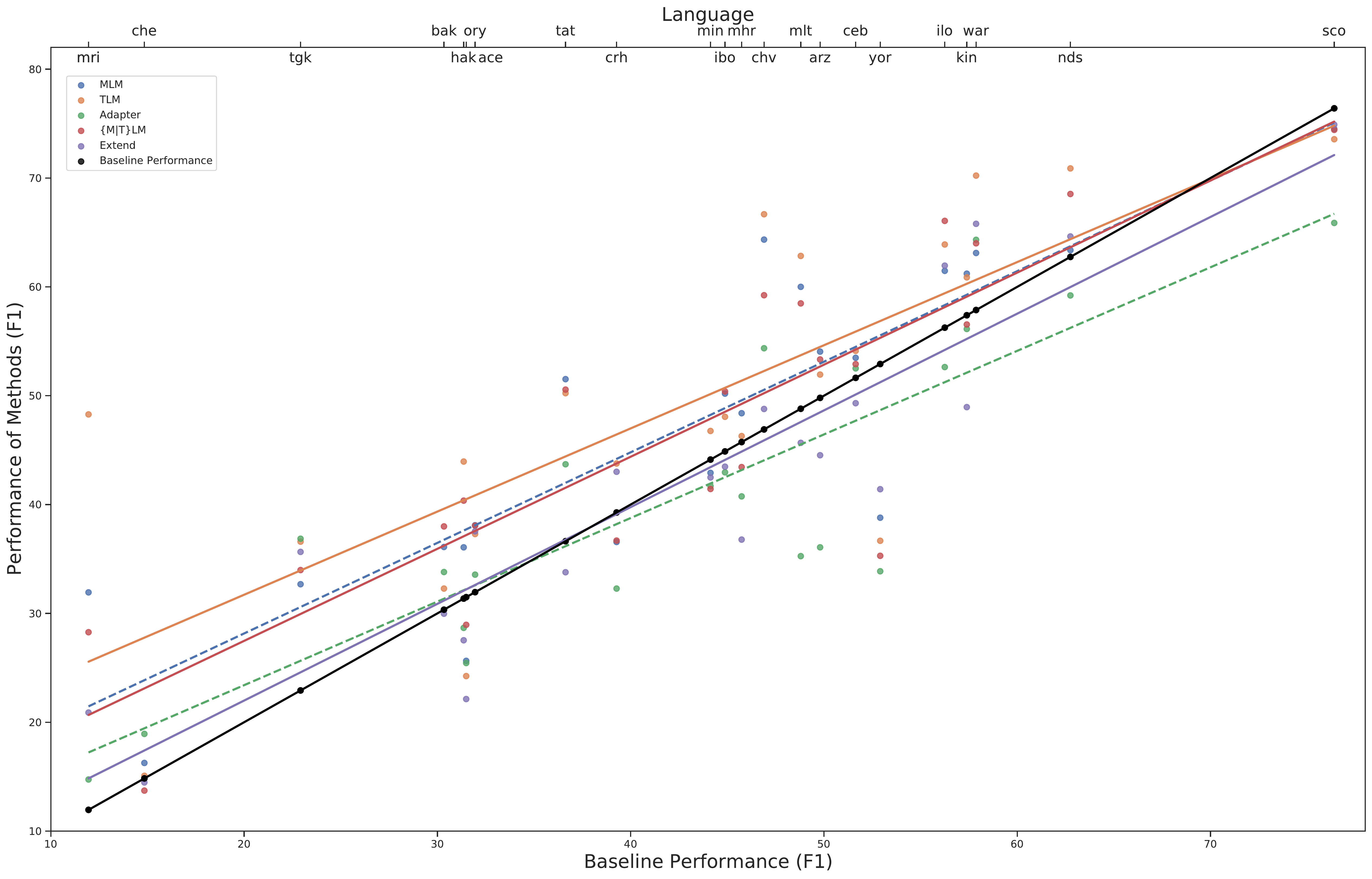}
    \caption{NER results (F1). Trendlines are created using linear regression.}
    \label{fig:ner_figure}
\end{figure*}
\begin{table}[ht]
    \setlength{\tabcolsep}{1pt}
    \centering
    \footnotesize
    \begin{adjustbox}{width=\columnwidth}
    \setlength\extrarowheight{2pt}

    \begin{tabular}{l|rrrrrr}
        \toprule
        \textbf{Lang.} & \textbf{XLM-R} & \textbf{+MLM} & \textbf{+TLM} & \textbf{+\{M$|$T\}LM} & \textbf{+Extend} & \textbf{+Adapters} \\
        \midrule
        ace & 31.95 & \textbf{38.10} & 37.29 & 38.06 & 37.54 & 33.56 \\
        arz & 49.80 & \textbf{54.05} & 51.94 & 53.33 & 44.53 & 36.07 \\
        bak & 30.34 & 36.10 & 32.28 & \textbf{37.99} & 29.97 & 33.80 \\
        ceb & 51.64 & 53.48 & \textbf{54.12} & 52.90 & 49.31 & 52.51 \\
        che & 14.84 & 16.26 & 15.08 & 13.72 & 14.47 & \textbf{18.93} \\
        chv & 46.90 & 64.34 & \textbf{66.67} & 59.23 & 48.78 & 54.36 \\
        crh & 39.27 & 36.56 & \textbf{43.77} & 36.69 & 43.01 & 32.28 \\
        hak & 31.36 & 36.07 & \textbf{43.95} & 40.36 & 27.53 & 28.67 \\
        ibo & 44.88 & 50.19 & 48.06 & \textbf{50.39} & 43.48 & 42.96 \\
        ilo & 56.25 & 61.47 & 63.89 & \textbf{66.06} & 61.95 & 52.63 \\
        kin & 57.39 & \textbf{61.21} & 60.87 & 56.54 & 48.95 & 56.13 \\
        mhr & 45.74 & \textbf{48.39} & 46.29 & 43.44 & 36.78 & 40.75 \\
        min & 44.13 & 42.91 & \textbf{46.76} & 41.43 & 42.49 & 41.70 \\
        mlt & 48.80 & 60.00 & \textbf{62.84} & 58.48 & 45.67 & 35.26 \\
        mri & 11.95 & 31.93 & \textbf{48.28} & 28.27 & 20.89 & 14.74 \\
        nds & 62.75 & 63.37 & \textbf{70.88} & 68.53 & 64.63 & 59.21 \\
        ory & \textbf{31.49} & 25.64 & 24.24 & 28.95 & 22.13 & 25.44 \\
        sco & \textbf{76.40} & 74.51 & 73.56 & 74.42 & 74.90 & 65.87 \\
        tat & 36.63 & \textbf{51.52} & 50.23 & 50.56 & 33.78 & 43.70 \\
        tgk & 22.92 & 32.68 & 36.59 & 33.98 & 35.65 & \textbf{36.86} \\
        war & 57.87 & 63.11 & \textbf{70.22} & 64.00 & 65.79 & 64.32 \\
        yor & \textbf{52.91} & 38.79 & 36.67 & 35.29 & 41.41 & 33.87 \\
        \midrule 
        Avg. & 43.01 & 47.30 & \textbf{49.29} & 46.94 & 42.44 & 41.07 \\
        $\Delta$ Avg. & 0.00 & 4.29 & \textbf{6.29} & 3.93 & -0.57 & -1.94 \\
        \bottomrule
    \end{tabular}
    \end{adjustbox}
    \caption{F1 score for all models on NER.}
    \label{tab:ner_result}
\end{table}
\section{Results}

\begin{figure*}[t]
    \centering
    \begin{adjustbox}{width=.9\textwidth}
    \includegraphics{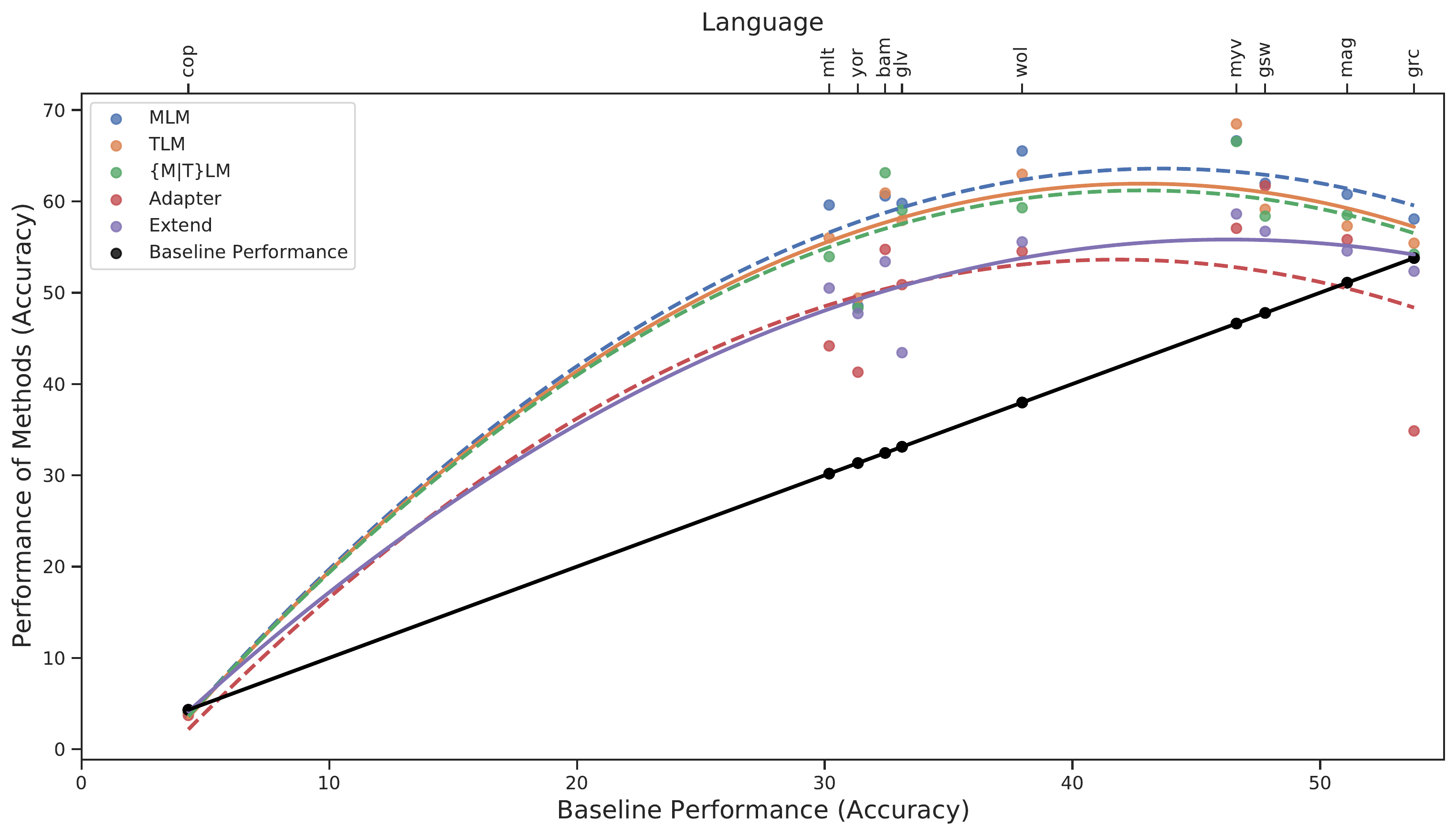} 
    \end{adjustbox}

    \caption{POS results (Accuracy). Trendlines are created by fitting a $2^{\text{nd}}$ order, least squares polynomial.}
    \label{fig:pos_fig}
\end{figure*}

We present results for NER and POS tagging in Tables \ref{tab:ner_result} and \ref{tab:pos_results}, respectively. We additionally provide plots of the methods' performances as compared to the XLM-R baseline in Figures \ref{fig:ner_figure} and \ref{fig:pos_fig}, showing performance trends for each model. 

\paragraph{NER}
We find that methods based on our most straightforward approach, continued pretraining (\texttt{+MLM}, \texttt{+TLM}, \texttt{+\{M|T\}LM}), perform best, with 3.93 to 6.29 F1 improvement over XLM-R. Both \texttt{+Extend} and \texttt{+Adapters} obtain a lower average F1 than the XLM-R baseline, which shows that they are not a good choice in our setting: either the size or the domain of the Bible causes them to perform poorly.
Focusing on the script of the target language (cf. Table \ref{tab:languages}), the average performance gain across all models is higher for Cyrillic languages than for Latin languages. Therefore, in relation to the source language script, performance gain is higher for target languages with a more distant script from the source.
When considering approaches which introduce new parameters, \texttt{+Extend} and \texttt{+Adapters}, performance only increases for Cyrillic languages and decreases for all others. However, when considering continued pretraining approaches, we find a performance increase for all scripts.

Looking at Figure \ref{fig:ner_figure}, we see that the lower the baseline F1, the larger the improvement of the adaption methods on downstream performance, with all methods increasing performance for the language for which the baseline is weakest. As baseline performance increases, the benefit provided by these methods diminishes, and all methods underperform the baseline for Scots, the language with the highest baseline performance.
We hypothesize that at this point the content of the Bible offers little to no extra knowledge for these languages compared to the existing knowledge in the pretraining data.

\paragraph{POS Tagging}
Our POS tagging results largely follow the same trend as those for NER, with continued pretraining methods achieving the highest increase in performance: between 15.81 and 17.61 points. Also following NER and as shown in Figure \ref{fig:pos_fig}, the largest performance gain can be seen for languages with a low baseline performance, and, as
the latter increases, the benefits obtained from adaptation become smaller.
However, unlike for NER, all methods show a net increase in performance, with \texttt{+Adapters}, the lowest performing adaptation model, achieving a gain of 9.01 points.
We hypothesize that a likely reason for this is the domain and style of the Bible. While it may be too restrictive to significantly boost downstream NER performance, it is still a linguistically rich resource for POS tagging, a task that is less sensitive to domain in general.

Additionally, there is a notable outlier language, Coptic, on which no model performs better than random choice (which corresponds to $6\%$ accuracy). This is because the script of this language is almost completely unseen to XLM-R, and practically all non-whitespace subwords map to the unknown token: of the 50\% of non-whitespace tokens, 95\% are unknown. While \texttt{+Extend} solves this issue, we believe that for a language with a completely unseen script the Bible is not enough to learn representations which can be used in a zero-shot setting.
\begin{table}[h]
    \centering

    \begin{adjustbox}{width=\columnwidth}
    \footnotesize
    \setlength\extrarowheight{2pt}
    \setlength{\tabcolsep}{1pt}

    \begin{tabular}{l|cccccc}
        \toprule
        \textbf{Lang.} & \textbf{XLM-R} & \textbf{+MLM} & \textbf{+TLM} & \textbf{+\{M$|$T\}LM}  & \textbf{+Extend} & \textbf{+Adapters} \\
        \midrule
        bam & 32.44 & 60.59 & 60.91 & \textbf{63.13}  & 53.40 & 54.74 \\
        cop & 4.31 & 4.02 & 4.03 & 4.03  & \textbf{4.35} & 3.70 \\
        glv & 33.12 & \textbf{59.78} & 57.91 & 59.05  & 43.43 & 50.88 \\
        grc & 53.79 & \textbf{58.07} & 55.42 & 54.21  & 52.35 & 34.86 \\
        gsw & 47.78 & \textbf{61.98} & 59.14 & 58.38  & 56.72 & 61.70 \\
        mag & 51.09 & \textbf{60.77} & 57.30 & 58.52  & 54.57 & 55.81 \\
        mlt & 30.18 & \textbf{59.60} & 56.00 & 53.95  & 50.50 & 44.17 \\
        myv & 46.62 & 66.63 & \textbf{68.48} & 66.55  & 58.62 & 57.05 \\
        wol & 37.97 & \textbf{65.52} & 62.97 & 59.30  & 55.56 & 54.52 \\
        yor & 31.34 & 48.54 & \textbf{49.40} & 48.31  & 47.71 & 41.29 \\
        \midrule
        Avg. & 36.86 & \textbf{54.55} & 53.16 & 52.54 & 47.72 & 45.875 \\
        $\Delta$ Avg. & 0.00 & \textbf{17.69} & 16.29 & 15.68 & 10.86 & 9.01 \\
    \bottomrule
    \end{tabular}
    \end{adjustbox}
    \caption{POS tagging accuracy.}
    \label{tab:pos_results}
\end{table}
\section{Case Study}

As previously stated, using the Bible as the corpus for adaptation is limiting in two ways: the extremely restricted domain as well as the small size. To separate the effects of these two aspects, we repeat our experiments with a different set of data. We sample sentences from the Wikipedia of each target language to simulate a corpus of similar size to the Bible which is not restricted to the Bible's domain or content. To further minimize the effect of domain, we focus solely on NER, such that the domain of the data is precisely that of the target task. Additionally, we seek to further investigate the effect on the downstream performance gains of these
adaptation methods when the source language is more similar to the target language. To this end, we focus our case study on three languages written in Cyrillic: Bashkir, Chechen, and Chuvash. We break up the case study into 3 settings, depending on the data used. In the first setting, we change the language of our labeled training data from English to Russian. While Russian is not necessarily similar to the target languages or mutually intelligible, we consider it to be more similar than English; Russian is written in the same script as the target languages, and there is a greater likelihood for lexical overlap and the existence of loanwords. In the second setting, we pretrain using Wikipedia and in the third setting we use both Wikipedia data as well as labeled Russian data.

To create our Wikipedia training data, we extract sentences with WikiExtractor \cite{Wikiextractor2015} and split them
with Moses SentenceSplitter \cite{koehn-etal-2007-moses}. To create a comparable training set for each language, we first calculate the total number of subword tokens found in the New Testament, and sample sentences from Wikipedia until we have an equivalent amount. In the setting where we use data from the New Testament and labeled Russian data, for \texttt{+TLM} and \texttt{+\{M|T\}LM} we additionally substitute the English Bible with the Russian Bible. When using Wikipedia, we omit results for \texttt{+TLM} and \texttt{+\{M|T\}LM}, as they rely on a parallel corpus.
\begin{table}[h]
    \setlength{\tabcolsep}{3.5pt}
    \centering
    \footnotesize

    \setlength\extrarowheight{0pt}

    \begin{tabular}{c|l|ccc|c}
        \toprule
         \textbf{Setting} & \multicolumn{1}{c|}{\textbf{Model}} & \textbf{bak} & \textbf{che} & \textbf{chv} & \textbf{Avg.}\\
         \midrule

            \multirow{2}{*}{\text{B-E}} & XLM-R & 30.34 & 14.84 & 46.90 & 30.69 \\
            & Best & 37.99 & 16.26 & 66.67 & 40.30 \\
            \midrule
             \multirow{6}{*}{\text{B-R}}& XLM-R & 53.84 & 43.94 & 51.16 & 49.65 \\
          &\hspace{4pt}+MLM & 58.46 & 38.67 & 55.09 & 50.74  \\
          &\hspace{4pt}+TLM & 53.58 & 27.78 & 50.96 & 44.10 \\
          &\hspace{4pt}+\{M$|$T\}LM & 57.99 & 31.64 & 57.14 & 48.92 \\
          &\hspace{4pt}+Extend & 39.86 & 27.16 & 46.32 & 37.78   \\
          &\hspace{4pt}+Adapters & 48.03 & 21.64 & 36.11 & 35.26  \\
          \midrule
          \multirow{3}{*}{\text{W-E}} &\hspace{4pt}+MLM & 43.12 & 18.18 & 74.82 & 45.37 \\
          &\hspace{4pt}+Extend & 30.26 & 20.93 & 35.62 & 28.94   \\
          &\hspace{4pt}+Adapters & 41.88 & 41.15 & 71.74 & 51.59  \\
          \midrule
          \multirow{3}{*}{\text{W-R}} &\hspace{4pt}+MLM & 61.19 & 55.89 & 67.42 & 61.50  \\
          &\hspace{4pt}+Extend & 36.84 & 44.08 & 35.46 & 38.79  \\
          &\hspace{4pt}+Adapters & 56.39 & 28.65 & 73.12 & 52.72  \\

         \bottomrule
    \end{tabular}

    \caption{Case study: Cyrillic NER (F1). \textit{Setting} describes the source of data for adaptation, either the (B)ible or (W)ikipedia, as well as the language of the finetuning data, (E)nglish or (R)ussian.}
    \label{tab:case_study}
\end{table}

\subsection{Results} We present the results of our case study in Table \ref{tab:case_study}. In the sections below, we refer to case study settings as they are described in the table caption.

\paragraph{Effects of the Finetuning Language} We find that  using Russian as the source language (the ``Russian baseline"; B-R w/ XLM-R) increases performance over the English baseline (B-E w/ XLM-R) by 18.96 F1. Interestingly, all of the adaptation methods utilizing the Bible do poorly in this setting (B-R), with \texttt{+MLM} only improving over the Russian baseline by 1.09 F1, and all other methods decreasing performance. We hypothesize that when adaptation data is limited in domain, as the source language approaches the target language in similarity, the language adaptation is mainly done in the finetuning step, and any performance gain from the unlabeled data is minimized.
This is supported by the previous NER results, where we find that, when using English as the source language, the adaptation methods lead to higher average performance gain over the baseline for Cyrillic languages, i.e., the more distant languages, as opposed to Latin languages. The adaptation methods show a larger improvement when switching to Wikipedia data (W-R), with \texttt{+MLM} improving performance by 11.85 F1 over the Russian baseline. Finally, the performance of \texttt{+Extend} when using Russian labeled data is similar on average regardless of the adaptation data (B-R, W-R), but noticeably improves over the setting which uses Wikipedia and English labeled data.

\paragraph{Effects of the Domain Used for Adaptation} Fixing English as the source language and changing the pretraining domain from the Bible to Wikipedia (W-E) yields strong improvements, with \texttt{+Adapters} improving over the English baseline by 20.9 F1 and \texttt{+MLM} improving by 14.68 F1. However, we note that, while the average of \texttt{+Adapters} is higher than that of \texttt{+MLM}, this is due to higher performance on only a single language. When compared to the best performing pretraining methods that use the Bible (B-E), these methods improve by 11.29 F1 and 5.30 F1 respectively. When using both Wikipedia and Russian data, we see the highest overall performance, and \texttt{+MLM} increases over the English baseline by 30.81 F1 and the Russian baseline by 11.85 F1.

\section{Limitations} One limitation of this work -- and other works which involve a high number of languages -- is task selection. While part-of-speech tagging and named entity recognition\footnote{We also note that the WikiANN labels are computer generated and may suffer from lower recall when compared to hand-annotated datasets.} are important, they are both low-level tasks largely based on sentence structure, with no requirement for higher levels of reasoning, unlike tasks such as question answering or natural language inference. While XTREME \cite{hu2020xtreme} is a great, diverse benchmark covering these higher level tasks, the number of languages is still limited to only 40 languages, all of which have Wikipedia data available. Extending these benchmarks to truly low resource languages by introducing datasets for these tasks will further motivate research on these languages, and provide a more comprehensive evaluation for their progress. 

Additionally, while the Bible is currently available in some form for 1611 languages, the available text for certain languages may be different in terms of quantity and quality from the Bible text we use in our experiments. Therefore, although we make no language-specific assumptions, our findings may not fully generalize to all 1611 languages due to these factors. Furthermore, this work focuses on analyzing the effects of adaptation methods for only a single multilingual transformer model. Although we make no model-specific assumptions in our methods, the set of unseen languages differs from model to model. Moreover, although we show improvements for the two tasks, we do not claim to have state-of-the-art results. In a low-resource setting, the best performance is often achieved through task-specific models. Similarly, translation-based approaches, as well as few-shot learning may offer additional benefits over a zero-shot setting. We also do not perform an extensive analysis of the target languages, or an analysis of the selected source language for finetuning. A better linguistic understanding of the languages in question would allow for a better selection of source language, as well as the ability to leverage linguistic features potentially leading to better results. 

Finally, by using a PMM, we inherit all
of that model's biases.
The biases captured by word embeddings are well known, and recent work has shown that contextual models are not free of biases either \cite{Caliskan_2017,kurita2019measuring}. The use of the Bible, and religious texts in general, may further introduce additional biases. Last, we acknowledge the environmental impact from the training of models on the scale of XLM-R \cite{strubell2019energy}.

\section{Conclusion} In this work, we evaluate the performance of continued pretraining, vocabulary extension, and adapters for unseen languages of XLM-R in a realistic low-resource setting. Using only the New Testament, we show that continued pretraining is the best performing adaptation approach, leading to gains of 6.29 F1 on NER and 17.69\% accuracy on POS tagging. We therefore conclude that the Bible can be a valuable resource for adapting PMMs to unseen languages, especially when no other data exists. Furthermore, we conduct a case study on three languages written in Cyrillic script. Changing the source language to one more similar to the target language reduces the effect of adaptation, but the performance of the adaptation methods relative to each other is preserved. Changing the domain of the adaptation data to one more similar to the target task while keeping its size constant improves performance.

\section*{Acknowledgments} We would like to thank the ACL reviewers for their constructive and insightful feedback as well as Yoshinari Fujinuma, Stéphane Aroca-Ouellette, and other members of the CU Boulder's NALA Group for their
advice and 
help.

\bibliographystyle{acl_natbib}
\bibliography{acl2021}
\clearpage
\appendix
\section{Appendix}
In Table \ref{tab:bible_counts}, we provide the number of subwords created by the XLM-R tokenizer from the New Testament of each target language, in addition to the specific version of the Bible we use, as found in the JHU Bible Corpus. In Table \ref{tab:app_ner_delta} and \ref{tab:app_pos_delta} we provide the relative performance of all adaptation methods as compared to baseline performance. 
\text{ } \\ \\ \\ \\
\begin{minipage}{\columnwidth}
    \small
    \centering
        \setlength\extrarowheight{2pt}
    \begin{adjustbox}{width=\columnwidth}
    
    \begin{tabular}{llr}
    \toprule
    Lang. & Bible Version & Bible Size (thousands) \\
    \bottomrule
         ace & ace-x-bible-ace-v1 & 536k \\
        arz & arz-x-bible-arz-v1 & 828k \\
        bak & bak-BAKIBT & 453k \\
        bam & bam-x-bible-bam-v1 & 429k \\
        ceb & ceb-x-bible-bugna2009-v1 & 384k \\
        che & che-x-bible-che-v1 & 523k \\
        chv & chv-CHVIBT & 519k \\
        cop & cop-x-bible-bohairic-v1 & 259k \\
        crh & crh-CRHIBT & 347k \\
        eng & eng-x-bible-kingjames-v1 & 461k \\
        glv & glv-x-bible-glv-v1 & 196k \\
        grc & grc-x-bible-textusreceptusVAR1-v1 & 322k \\
        gsw & gsw-x-bible-alemannisch-v1 & 351k \\
        hak & hak-x-bible-hak-v1 & 598k \\
        ibo & ibo-x-bible-ibo-v1 & 458k \\
        ilo & ilo-x-bible-ilo-v1 & 378k \\
        kin & kin-x-bible-bird-youversion-v1 & 344k \\
        mag & mag-MAGSSI & 388k \\
        mhr & mhr-x-bible-mhr-v1 & 398k \\
        min & min-x-bible-min-v1 & 505k \\
        mlt & mlt-x-bible-mlt-v1 & 389k \\
        mri & mri-x-bible-mri-v1 & 411k \\
        myv & myv-x-bible-myv-v1 & 463k \\
        nds & nds-x-bible-nds-v1 & 333k \\
        ory & ory-x-bible-ory-v1 & 386k \\
        rus & rus-x-bible-kulakov-v1 & 283k \\
        sco & sco-x-bible-sco-v1 & 30k \\
        tat & tat-TTRIBT & 438k \\
        tgk & tgk-TGKIBT & 233k \\
        war & war-x-bible-war-v1 & 401k \\
        wol & wol-x-bible-wol-v1 & 383k \\
        yor & yor-x-bible-yor-v1 & 450k \\
        \midrule
        Avg.  &   & 402k \\
    \bottomrule
    \end{tabular}
    \end{adjustbox}
    \captionof{table}{Size of the New Testaments of each language, along with the specific Bible version. Size is calculated in subword units using the base XLM-Roberta tokenizer.}
    \label{tab:bible_counts}
\end{minipage}

\onecolumn

\begin{minipage}{\columnwidth}
    \centering

    \small
    \setlength\extrarowheight{2pt}
    \setlength{\tabcolsep}{1pt}
    \resizebox{0.5\columnwidth}{!}{%
    \begin{tabular}{l|cccccc}
        \toprule
        \textbf{Lang.} & \textbf{XLM-R} & $\Delta$\textbf{MLM} & $\Delta$\textbf{TLM} & $\Delta$\textbf{\{M$|$T\}LM}  & $\Delta$\textbf{Extend} & $\Delta$\textbf{Adapters} \\
        \midrule 
        bam & 32.44 & 28.15 & 28.47 & 30.69  & 20.96 & 22.30 \\
        cop & 4.31 & -0.29 & -0.28 & -0.28  & 0.04 & -0.61 \\
        glv & 33.12 & 26.66 & 24.79 & 25.93  & 10.31 & 17.76 \\
        grc & 53.79 & 4.28 & 1.63 & 0.42  & -1.44 & -18.93 \\
        gsw & 47.78 & 14.20 & 11.36 & 10.60  & 8.94 & 13.92 \\
        mag & 51.09 & 9.68 & 6.21 & 7.43  & 3.48 & 4.72 \\
        mlt & 30.18 & 29.42 & 25.82 & 23.77  & 20.32 & 13.99 \\
        myv & 46.62 & 20.01 & 21.86 & 19.93  & 12.00 & 10.43 \\
        wol & 37.97 & 27.55 & 25.00 & 21.33  & 17.59 & 16.55 \\
        yor & 31.34 & 17.20 & 18.06 & 16.97  & 16.37 & 9.95 \\
        \midrule
        Avg. & 36.86 & 17.69 & 16.29 & 15.68  & 10.86 & 9.01 \\
    \bottomrule
    \end{tabular}}

    \captionof{table}{Accuracy deltas for POS tagging compared to baseline}
    \label{tab:app_pos_delta}
\end{minipage}
\text{ } \\ \\ \\ \\ \\ \\ 
\begin{minipage}{0.5\columnwidth}
    \setlength{\tabcolsep}{1pt}
    \centering
    \small
    \begin{adjustbox}{width=\columnwidth}
    \setlength\extrarowheight{2pt}

    \begin{tabular}{l|cccccc}
        \toprule
        \textbf{Lang.} & \textbf{XLM-R} & $\Delta$\textbf{MLM} & $\Delta$\textbf{TLM} & $\Delta$\textbf{\{M$|$T\}LM}  & $\Delta$\textbf{Extend} & $\Delta$\textbf{Adapters} \\
        \midrule 
        ace & 31.95 & 6.15 & 5.34 & 6.11  & 5.59 & 1.61 \\
        arz & 49.80 & 4.25 & 2.14 & 3.53  & -5.27 & -13.73 \\
        bak & 30.34 & 5.76 & 1.94 & 7.65  & -0.37 & 3.46 \\
        ceb & 51.64 & 1.84 & 2.48 & 1.26  & -2.33 & 0.87 \\
        che & 14.84 & 1.42 & 0.24 & -1.12  & -0.37 & 4.09 \\
        chv & 46.90 & 17.44 & 19.77 & 12.33  & 1.88 & 7.46 \\
        crh & 39.27 & -2.71 & 4.50 & -2.58  & 3.74 & -6.99 \\
        hak & 31.36 & 4.71 & 12.59 & 9.00  & -3.83 & -2.69 \\
        ibo & 44.88 & 5.31 & 3.18 & 5.51  & -1.40 & -1.92 \\
        ilo & 56.25 & 5.22 & 7.64 & 9.81  & 5.70 & -3.62 \\
        kin & 57.39 & 3.82 & 3.48 & -0.85  & -8.44 & -1.26 \\
        mhr & 45.74 & 2.65 & 0.55 & -2.30  & -8.96 & -4.99 \\
        min & 44.13 & -1.22 & 2.63 & -2.70  & -1.64 & -2.43 \\
        mlt & 48.80 & 11.20 & 14.04 & 9.68  & -3.13 & -13.54 \\
        mri & 11.95 & 19.98 & 36.33 & 16.32  & 8.94 & 2.79 \\
        nds & 62.75 & 0.62 & 8.13 & 5.78  & 1.88 & -3.54 \\
        ory & 31.49 & -5.85 & -7.25 & -2.54  & -9.36 & -6.05 \\
        sco & 76.40 & -1.89 & -2.84 & -1.98  & -1.50 & -10.53 \\
        tat & 36.63 & 14.89 & 13.60 & 13.93  & -2.85 & 7.07 \\
        tgk & 22.92 & 9.76 & 13.67 & 11.06  & 12.73 & 13.94 \\
        war & 57.87 & 5.24 & 12.35 & 6.13  & 7.92 & 6.45 \\
        yor & 52.91 & -14.12 & -16.24 & -17.62  & -11.50 & -19.04 \\
        \midrule
        Avg. & 43.01  & 4.29 & 6.29 & 3.93  & -0.57 & -1.94 \\
        \bottomrule
    \end{tabular}
    \end{adjustbox}
    \captionof{table}{F1 deltas for NER compared to baseline}
    \label{tab:app_ner_delta}
\end{minipage}

\end{document}